\documentclass[preprint,times,review,1p,10pt]{elsarticle}

\usepackage{amssymb}
\usepackage{amsmath}
\usepackage{booktabs}
\usepackage{multirow}

\usepackage{amssymb} 
\usepackage{amsfonts}
\usepackage{algorithm}
\usepackage{algorithmic}
\usepackage{bm}
\usepackage{multirow}
\usepackage{xcolor}
\usepackage{dsfont}
\usepackage{amssymb}
\usepackage{amsmath}
\usepackage{float}
\usepackage{subfig}
\usepackage{graphicx}
\usepackage{algorithm}
\usepackage{algorithmic}
\usepackage{ulem}
\usepackage{url}
\usepackage{booktabs}
\usepackage{multicol}
\usepackage{multirow}
\usepackage{wrapfig}
\usepackage{hyperref}
\usepackage{amsmath}
\usepackage{blkarray}
\usepackage{bm}

\newcommand{\Indicator}{\mathds{1}}

\journal{Pattern Recognition}

\begin{document}

\begin{frontmatter}



\title{Learning Coherent Matrixized Representation in Latent Space for 
Volumetric 4D Generation}


\author[add1]{Qitong Yang}
\author[add1]{Mingtao Feng}
\author[add1]{Zijie Wu}
\author[add2]{Shijie Sun}
\author[add1]{Weisheng Dong}
\author[add3]{Yaonan Wang}
\author[add4]{Ajmal Mian}


\affiliation[add1]{organization={School of Artificial Intelligence},
            addressline={Xidian University}, 
            city={Xi’an},
            postcode={710126}, 
            state={Shanxi},
            country={China}}

\affiliation[add2]{organization={School of Information Engineering},
            addressline={Chang'an University}, 
            city={Xi’an},
            postcode={710064}, 
            state={Shanxi},
            country={China}}

\affiliation[add3]{organization={School of Electrical and Information Engineering},
            addressline={Hunan University}, 
            city={Changsha},
            postcode={410082}, 
            state={Hunan},
            country={China}}

\affiliation[add4]{organization={Computer Science and Software Engineering},
            addressline={The University of Western Australia}, 
            city={Perth},
            postcode={6009}, 
            state={WA},
            country={Australia}}            


\begin{abstract}
    Directly learning to model 4D content, including shape, color, and motion, is challenging. Existing methods rely on pose priors for motion control, resulting in limited motion diversity and continuity in details. To address this, we propose a framework that generates volumetric 4D sequences, where 3D shapes are animated under given conditions (text-image guidance) with dynamic evolution in shape and color across spatial and temporal dimensions, allowing for free navigation and rendering from any direction. We first use a coherent 3D shape and color modeling to encode the shape and color of each detailed 3D geometry frame into a latent space. Then we propose a matrixized 4D sequence representation allowing efficient diffusion model operation. Finally, we introduce spatio-temporal diffusion for 4D volumetric generation under given images and text prompts. Extensive experiments on the ShapeNet, 3DBiCar, DeformingThings4D and Objaverse datasets for several tasks demonstrate that our method effectively learns to generate high quality 3D shapes with consistent color and coherent mesh animations, improving over the current methods. Our code will be publicly available.
\end{abstract}




\end{frontmatter}




\section{Introduction}
Generative modeling of dynamic 3D scenes can potentially revolutionize how we create animations, games, movies, simulations, and entire virtual worlds. Encouraging progress has been made in synthesizing diverse 3D objects via generative models. However, current methods typically synthesize static 3D scenes, and the quality and diversity of the generated shapes still require improvement. While image diffusion models~\cite{sheynin2022knn} have been successfully extended to video generation~\cite{jain2024video}, few studies have similarly extended 3D synthesis~\cite{tang2023dreamgaussian} to 4D generation by adding a temporal dimension.

\begin{figure}[t]
    \centering    \includegraphics[width=0.8\linewidth]{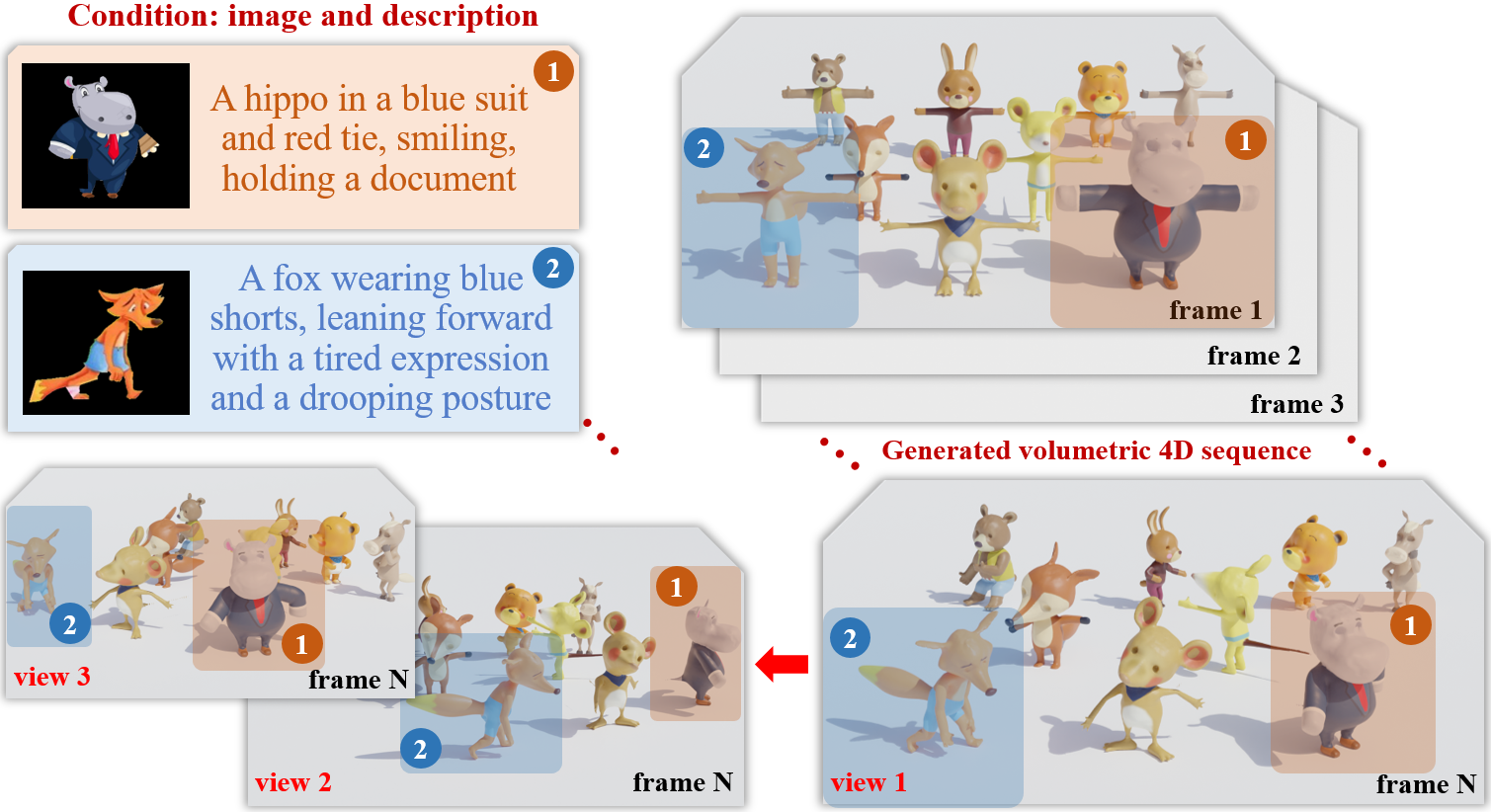}
    \vspace{-2mm}
    \caption{Proposed image-text conditioned 4D generation with high 3D shape quality, color fidelity and sequence coherence, each frame enables free navigation and rendering from any direction.}
    \label{fig:intro}
    \vspace{-5mm}
\end{figure}

Recent works~\cite{erkocc2023hyperdiffusion,ren2023dreamgaussian4d} combine the controllability of dynamic 3D objects with the expressivity of emerging diffusion models. 4D representations are proposed to construct an underlying representation of dynamic 3D objects, allowing the diffusion model to learn the distribution of 3D synthesis with an additional temporal dimension. In general, 4D representations can be categorized into prior-based and free-form methods depending on the 3D representation of the output shape. Prior-based methods are mostly derived from shape parameters combined with a series of skeletal pose~\cite{DING201875} parameters to model dynamic sequences, e.g., 4D human motion generation~\cite{KRAWCZYK2023109444}. Although they produce plausible results, the generated skeleton human motions assuming a human model prior such as SMPL~\cite{SMPL}, lacks motion diversity and only supports limited time span. 
Free-form methods leveraging dynamic Neural Radiance Fields (NeRF) combine the benefits of video and 3D generative models~\cite{jiang2023consistent4d}, and require heavy computations to generate a 4D NeRF. Moreover, their generated motions are not well-controllable. The 4D Gaussian splatting-based free-form method~\cite{JIANG2025111426, ren2023dreamgaussian4d} can produce high-quality renderings, but artifacts often appear when the view changes. This is due to the lack of geometric priors in 4D Gaussian splatting-based free-form approaches. In addition, free-form methods often produce low-quality rendering results during arbitrary navigation. This issue arises because the generated 4D content lacks the essential volumetric properties needed for accurate representation.
To this end, \textit{we pose a question,} how can we efficiently represent a 4D sequence to generate with high-quality geometry, sequence consistency, and alignment with input guidance, while maintaining diversity and allowing for free navigation and rendering?

To address these challenges, we propose several techniques to achieve volumetric 4D sequence generation under given conditions, as shown in Fig.~\ref{fig:intro}. We \textit{first} design a coherent 3D shape and color modeling method to encode the shape and color information of each frame. To ensure consistent color representation during shape deformation, we propose a continuous color field that constructs a continuous function to represent color values within the geometric space, which is represented by Signed Distance Function (SDF). We decouple shape and color, and design a decoder-only structure to map the shape and color information into two latent vectors. Our coherent 3D shape and color modeling links SDF with color while decoupling color, allowing for independently manipulating color without affecting the shape. Our method does not require an off-the-shelf 2D diffusion model like texture rendering based colored 3D object generation~\cite{cheng2023sdfusion}.
\textit{Next}, since direct application of diffusion models to the volumetric SDF and color is computationally intensive, we introduce a  matrixized 4D sequence representation, extending the 3D shape and coherent color representations along the temporal dimension. The representations across all frames are jointly concatenated to a matrix capturing the 4D distribution of the sequence. The 4D representation allows the diffusion model to efficiently operate on the temporal dimension, resulting in the generation of intricate 4D sequences with high fidelity.
\textit{Thirdly}, equipped with the efficient 4D representation, we propose an image-text conditioned latent diffusion model for 4D sequence generation. This model captures the intra-frame information, as well as frame-to-frame and frame-to-global relationships, using the proposed hierarchical conditional spatio-temporal attention mechanism to enforce temporal consistency on the content during generation. 

Our main contributions are: (1) We propose an image-text conditioned 4D volumetric generation framework that enables free navigation with high-quality, consistent rendering. 
(2) We present a coherent 3D shape and color modeling strategy that ensures consistent color representation during deformation via a continuous color field within geometric space, while enabling flexible shape-color combinations. (3) We introduce a matrixized 4D representation method, independent of pre-rigged pose priors, that facilitates efficient correlation learning along the temporal dimension. (4) We design a hierarchical conditional spatio-temporal attention mechanism to enforce temporal consistency in the generated content during the diffusion process. We assess unconditioned/conditioned 4D sequences generation across multiple datasets, to show that our approach achieves promising generation performances.

\section{Related Work}
\vspace{-2mm}
\noindent \textbf{3D Generative Models}  have been extensively explored based on point clouds~\cite{wu2023sketch}, voxel grids~\cite{lin2023infinicity}, meshes~\cite{zhang2021sketch2model}, and SDFs~\cite{cheng2023sdfusion}. 
Beyond the generation of 3D shapes, efforts have been made in generating colored objects \cite{cheng2023sdfusion, luo2023rabit}. Most methods are based on texture rendering, NeRF, score distillation, or Gaussian splatting. Recent works~\cite{erkocc2023hyperdiffusion,ren2023dreamgaussian4d} combine the controllability of colored dynamic 3D objects with the expressivity of emerging diffusion models. However, their downsides are long optimization time and poor performance beyond the 
training views.
Arguably, these methods are ineffective for applications involving dynamic signals, e.g. colored 3D moving objects, since the temporal information is not captured.

\noindent \textbf{4D Representations} have primarily taken two directions. One approach treats 4D scenes as functions of spatial dimensions (x, y, z) extended by the temporal dimension (t) or latent codes \cite{wu20234d, li2022neural}. The alternative approach involves integrating deformation fields with static, canonical 3D models \cite{park2021nerfies, MIAO2024110729}. A key challenge in 4D representation is maintaining computational efficiency along with temporal consistency.
Explicit and hybrid representations yield notable improvements in speed and reconstruction quality, e.g., planar decomposition for 4D space-time grids ~\cite{HUANG2024110758}, hash representations~\cite{turki2023suds}, and other innovative structures~\cite{abou2024particlenerf} have shown promise. 
However, current research lacks methods for coherent shape and color representation in 4D spatio-temporal space. While recent advances in volumetric video~\cite{jiang2024robust} show promising results, the need for extensive manual stabilization of mesh sequences and the generation of large assets in existing workflows hinders direct volume-based 4D content generation.

\noindent \textbf{4D Generative Models.} A prominent research direction employs text-to-video diffusion models to refine 4D representations, exemplified by the optimization of structures like Hexplane~\cite{cao2023hexplane} or K-plane ~\cite{fridovich2023k}. This process typically involves crafting camera trajectories and applying score distillation sampling on the rendered video sequences. Method~\cite{4d-fy} focuses on enhancing the photorealism of these representations. However, the motion elements are tightly coupled with the 3D content, resulting in lack of diversity and control over the generated motions. An alternate approach uses priors, mostly derived from the skeleton shape parameters and a series of pose parameters to model dynamic sequences~\cite{DING201875}. Their motion representations, while plausible, lack compactness and are limited to short time spans. Our method supports the generation of variable-length 4D sequences while allowing for the free combination of shape, motion, and color.

\begin{figure*}[t]
    \centering
    \includegraphics[width=\linewidth]{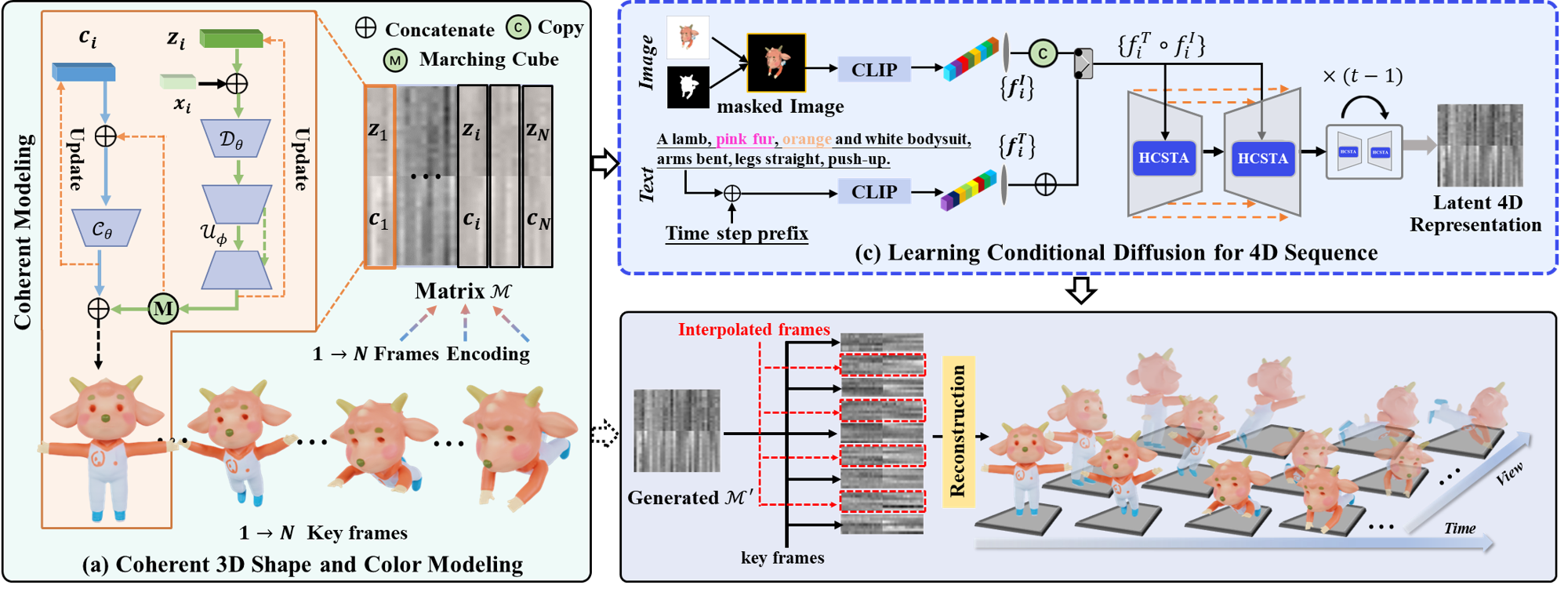}
    \vspace{-7mm}
    \caption{Method overview. The shape and color latent vectors for full sequences are jointly concatenated into a matrixized 4D latent representation $\mathcal{M}$. The input masked image and text are encoded via CLIP~\cite{CLIP} to condition the diffusion process of  $\mathcal{M}$. The volumetric 4D sequences are then reconstructed from the generated$\mathcal{M}'$, with latent frame interpolation enabling variable-length sequence generation.}
    \label{fig:model}
    \vspace{-3mm}
\end{figure*}

\section{Method}

Our objective is to generate volumetric 4D sequence for an object which can provide plausible geometry and appearance at specified time, enabling freely navigate and rendering. Fig.~\ref{fig:model} shows an overview of our method. We first introduce a coherent 3D shape and color modeling, which can handle the non-rigid deformation of shape and color details in the 4D sequence. Additionally, we propose a latent matrixized 4D sequence representation that compresses 4D sequence thus enabling generation. Finally we introduce a multi-condition diffusion model with proposed Hierarchical Conditional Spatio-Temporal Attention (HCSTA) block to generate the latent matrixized 4D sequence representation under the given conditions. Our method achieves high generation quality while providing exact object at specified times and significantly reduces the occurrence of artifacts compared to other free-form methods, due to its volumetric generation results.

\label{sec:implicit-representation}
\subsection{Coherent 3D Shape and Color Modeling}
\begin{figure}[t]
    \centering    \includegraphics[width=\linewidth]{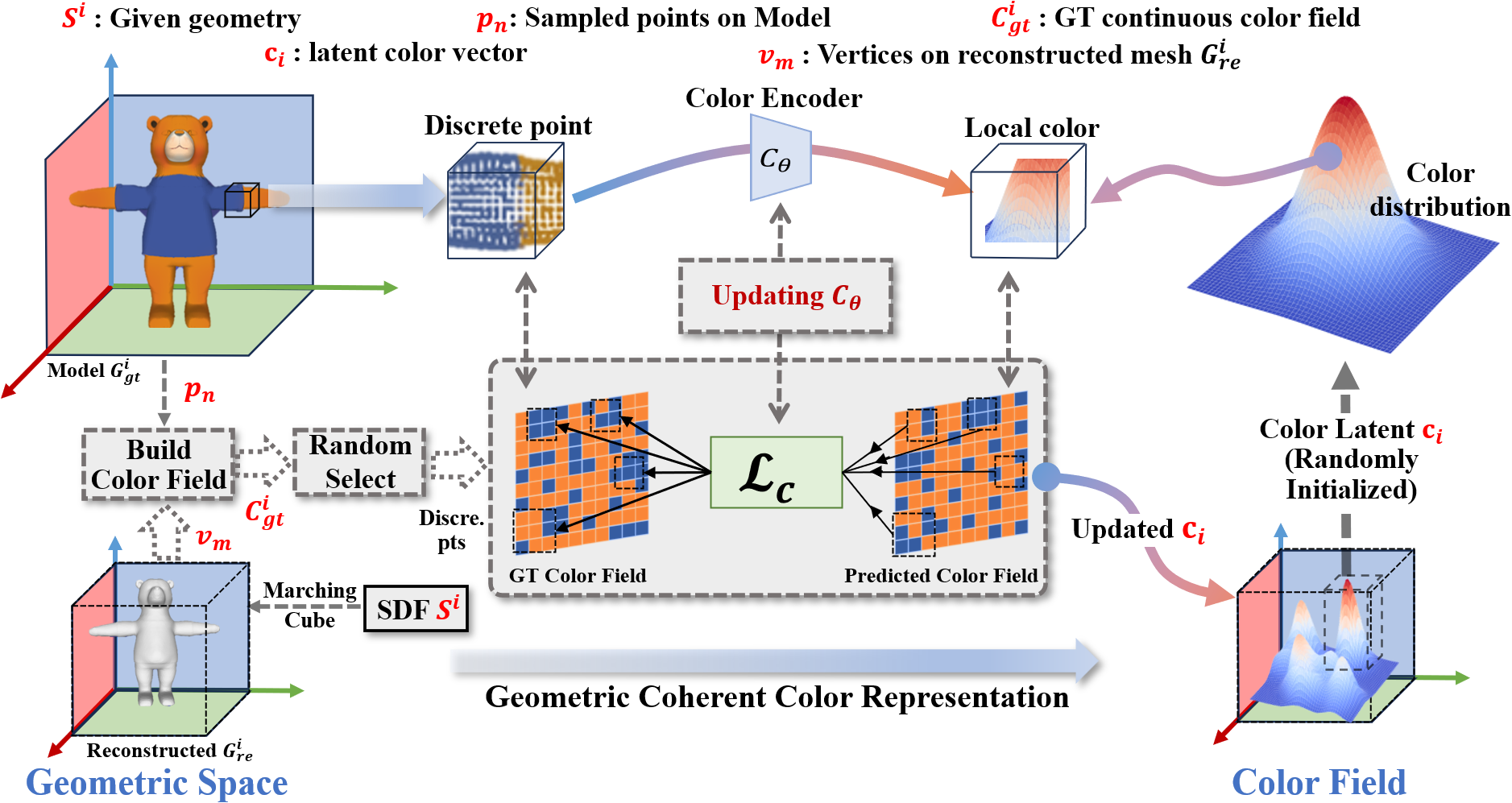}
    \vspace{-6mm}
    \caption{Pipeline for learning geometric coherent color representation. During each training epoch, a subset of vertices is randomly selected.}
    \label{fig:color-learning}
    \vspace{-3mm}
\end{figure}

The goal of the coherent 3D shape and color modeling is to create an efficient representation of a colored 3D object, and enables the foundation for representing volumetric 4D data. Two elements of the 3D representation are crucial for shape quality and play an important role in the learning of the subsequent 4D stage: (1) the
representation should be able to handle the non-rigid deformation of shape and color details in the 4D stage and (2) provide the possibility to learn spatio-temporal coherence at the 4D stage. To achieve these goals, we decompose the shape and color, proposing to use both shape latent representation and color latent representation for coherent 3D shape and color modeling, as shown in Fig.~\ref{fig:color-learning}.

\noindent\textbf{Fine-grained 3D Shape Representation.}
We represent 3D geometry as a continuous SDF using a shape latent vector $\mathbf{z}_i\in \mathbb{R}^{d_1}$, enabling arbitrary geometry modeling with varying topology. We train an auto-decoder $D_{\theta}$ to generate a coarse SDF $\mathcal{S}_\mathrm{coarse}^i$, which is further refined by a 3D U-Net $U_\phi$ to produce the final SDF $\mathcal{S}^i$:
\begin{equation}
    \mathcal{S}_\mathrm{coarse}^i = D_\theta(\mathbf{z}_i, x_i), \qquad \mathcal{S}^i = U_\phi(\mathcal{S}_\mathrm{coarse}^i),
\end{equation}
where $x_i$ is the query coordinate of the 3D shape.
During training, we optimize the joint log posterior over all shapes, with the loss function $\mathcal{L}_s$ defined as:
\begin{equation}
   \mathcal{L}_s = L_1(\mathcal{S}^i, \mathcal{S}_\mathrm{gt}^i)+\frac{1}{\sigma^2}\|\mathbf{z}_i\|_2^2.
\end{equation}
During inference, the shape latent code $\hat{\mathbf{z}_i}$ for a target SDF $\mathcal{S}^i_{gt}$ is estimated via Maximum-a-Posterior (MAP) as:
\begin{equation}
    \hat{\mathbf{z}_i} =  \mathop{\arg\min}_{\mathbf{z}} \displaystyle \sum^{N}_{i=1}\mathcal{L}(U_\phi(D_\theta(\mathbf{z}_i, x_i)), \mathcal{S}_\mathrm{gt}^i)+\frac{1}{\sigma^2}\|\mathbf{z}_i\|_2^2.
\end{equation}

\noindent\textbf{Geometric Coherent Color Representation.}
We then propose a continuous color field within the geometric space to ensure consistent color representation during deformation. Given the predicted $\mathcal{S}^i$, we first perform marching cube to reconstruct shape $G_{re}^i$ and then use a remesh method~\cite{khan2020surface} to enhance color details by increasing the vertices. As shown in Fig.~\ref{fig:color-learning}, we align these vertices $v = \{v_m\}|^M_{m=1}$ and use a KD-tree to extract color values $\mathbf{C}^i$ from the ground truth $G_\mathrm{gt}^i$. These operations leverage the information obtained from the
learned shape, indicating that the construction of color representation is constrained by the shape. To ensure that the continuous color field fully represents the color distribution of the object and is generalizable to a broader range of geometries, avoiding fixing the combination of shape and color, we randomly sample points $p_n$ both inside and outside the reconstructed surface and use an interpolation function to compute the color values of these points. We construct the continuous color field by: $\mathbf{C}^i_\mathrm{gt} = \{\Phi_\theta(\mathbf{C}^i, v_m, p_n)| n = [1,2,\cdots,N]\}$, where $\Phi_\theta$ is the interpolation function. Respectively, we use another latent vector $\mathbf{c}_i\in \mathbb{R}^{d_2}$ to represent the continuous color field $\mathbf{C}^i_\mathrm{gt}$ of a 3D object as:
\begin{equation}
    \mathbf{C}^i_\mathrm{pred} = C_\theta(\mathbf{c}_i, p_n),
\end{equation}
where $p_n$ are the random points, $\mathbf{C}^i_\mathrm{pred}$ is the predicted continuous color field.
The color decoder $C_\theta$ and color latent vector $\mathbf{c}_i$ are optimized using a combination of L1-norm and structural similarity loss:
\begin{small}
    \begin{equation}\begin{aligned}
    \mathcal{L}_c =\lambda_1(L_1(\mathbf{C}^i_\mathrm{pred},   \mathbf{C}^i_\mathrm{gt}) + \frac{1}{\sigma^2}\|\mathbf{c}_i\|_2^2)) 
    + \lambda_2L_\mathrm{SSIM}(\mathbf{C}^i_\mathrm{pred}, \mathbf{C}^i_\mathrm{gt}).
\end{aligned}\end{equation}
\end{small}
During inference, the color latent code $\hat{\mathbf{c}_i}$ is acquired using MAP:
\begin{equation}
    \hat{\mathbf{c}_i} =  \mathop{\arg\min}_{\mathbf{c}} \displaystyle \sum^{N}_{i=1}\mathcal{L}_c(C_\theta(\mathbf{c}_i, p_i), \mathbf{C}^i_\mathrm{gt})+\frac{1}{\sigma^2}\|\mathbf{c}_i\|_2^2.
\end{equation}

Note that the color filed links the color value with the SDF value because we use random points $p_k$ near the reconstructed vertices to construct the continuous color field. Additionally, our decoupled design for shape and color latent representation enables independently manipulating color without affecting the shape.
Our method synergistically maps shape and color to latent representations to ensure coherent and detailed 3D representations, while decoupling the shape and color representations, allowing for flexible combinations of shapes and colors. This approach facilitates smooth non-rigid deformations and maintains color consistency, thereby providing a robust foundation for further spatio-temporal learning and generation tasks.

\label{sec:4d-representation}
\subsection{Matrixized Representations in Latent Space}
Our goal is to construct a 4D representation optimized for generation. The most straightforward approach is to aggregate $N$ spatial-temporal elements frame-by-frame, leading to time complexity $O(T^2 N^2)$ for a sequence with $T$ frames and $N$ elements each, which is computationally infeasible. A simplified approach is predicting offsets from a single global code~\cite{tang2021learning, niemeyer2019occupancy, lei2022cadex} but it loses significant surface geometry, color, and temporal evolution information. The 4D sequence should exhibit smoothness both spatially and temporally, considering the limited elasticity and velocity of object. Therefore, a continuous latent sets of shape and color is responsible for representing a volumetric 4D sequence. Building upon our Coherent 3D Shape and Color Modeling, our latent representations can form a meaningful continuous latent space when training data is sufficient~\cite{rombach2022high, park2019deepsdf}, exhibiting spatiotemporal continuity. Based on spatiotemporal continuity, we propose a keyframe-based Matrixized 4D Representation.

We treat our latent 4D sequence representation as frame stamps on the volumetric 4D sequence, which takes a serial of keyframes for observation. 
Based on our interpolatable and meaningful continuous latent space, we propose a $K$-controlled matrixized 4D representation:
\begin{equation}
\mathcal{M} = \begin{bmatrix}
\mathbf{z}_1 & \overbrace{[\mathbf{z}_{u1\xrightarrow{}\cdots}]}^{K\text{ frames}}  & \mathbf{z}_2 & \overbrace{[\mathbf{z}_{u2\xrightarrow{}\cdots}]}^{K\text{ frames}}&\cdots &\overbrace{[\mathbf{z}_{un-1\xrightarrow{}\cdots}]}^{K\text{ frames}} & \mathbf{z}_T \\ 
\mathbf{c}_1 & \underbrace{[\mathbf{c}_{u1\xrightarrow{}\cdots}]}_{K\text{ frames}} & \mathbf{c}_2 & \underbrace{[\mathbf{c}_{u2\xrightarrow{}\cdots}]}_{K\text{ frames}} &\cdots &\underbrace{[\mathbf{c}_{un-1\xrightarrow{}\cdots}]}_{K\text{ frames}} & \mathbf{c}_T \,
\end{bmatrix},
\end{equation}
where $\mathbf{z}_i$, $\mathbf{c}_i$ are shape and color latent vectors for keyframe $i$, $[\mathbf{z}_{ui \rightarrow \cdots}]$, $[\mathbf{c}_{ui \rightarrow \cdots}]$ denote the unseen $K$ frames in the proposed matrixized representation, which can be recovered via interpolation from adjacent frames to balance efficiency and accuracy.
We determine $K$ by evaluating the quality of unseen frames using the reconstruction loss defined in our coherent modeling on the interpolated 4D sequences:
\begin{align}
\arg&\max\limits_{K}\mathbb{E}\left[\mathcal{L}_s(\psi(\mathbf{z}_1\cdots\mathbf{z}_T))+ \mathcal{L}_c(\psi(\mathbf{c}_1\cdots\mathbf{c}_T))-\mathcal{L}_s(\hat{\mathbf{z}}_1\cdots\hat{\mathbf{z}}_T)-\mathcal{L}_c(\hat{\mathbf{c}}_1\cdots\hat{\mathbf{c}}_T)\right]. \\
&\text{s.t.} \mathbb{E}\left[\mathcal{L}_s(\psi(\mathbf{z}_1\cdots\mathbf{z}_T))+ \mathcal{L}_c(\psi(\mathbf{c}_1\cdots\mathbf{c}_T))-\mathcal{L}_s(\hat{\mathbf{z}}_1\cdots\hat{\mathbf{z}}_T)-\mathcal{L}_c(\hat{\mathbf{c}}_1\cdots\hat{\mathbf{c}}_T)\right] \leq \sigma
\label{eq:linear_intepl}
\end{align}                        

Here, $(\hat{\mathbf{z}}_1\cdots\hat{\mathbf{z}}_T)$ and $(\hat{\mathbf{c}}_1\cdots\hat{\mathbf{c}}_T)$ are the original latents of 4D sequence, $\sigma$ is a small value computed as a proportion of $\mathbb{E}\left[\mathcal{L}_s(\hat{\mathbf{z}}_1, \cdots, \hat{\mathbf{z}}_T) + \mathcal{L}_c(\hat{\mathbf{c}}_1, \cdots, \hat{\mathbf{c}}_T)\right]$, $\psi$ denotes an interpolation function for recovering $K$ unseen frames, which can be implemented using linear interpolation
\begin{equation}
    \begin{bmatrix}[\mathbf{z}_{ui\xrightarrow{}\cdots}] \\ [\mathbf{c}_{ui\xrightarrow{}\cdots}]\end{bmatrix} = \Lambda^{T}\begin{bmatrix}\mathbf{z}_i \\ \mathbf{c}_i \,\end{bmatrix} + (1-\Lambda^{T})\begin{bmatrix}\mathbf{z}_{i+1} \\ \mathbf{c}_{i+1} \,\end{bmatrix},
\label{eq:map_intepl}
\end{equation}
through maximum a posterior (MAP) estimation:
\begin{equation}
    \min \mathbb{E} \left[ -\sum \log p([\mathbf{z}_{ui\xrightarrow{}\cdots}], [\mathbf{c}_{ui\xrightarrow{}\cdots}] \mid \mathbf{z}_{i}, \mathbf{c}_{i}, \mathbf{z}_{i+1}, \mathbf{c}_{i+1}) \right].
\end{equation}
Notably, this latent 4D sequence representation reduces the complexity to $O(\frac{T^2}{K^2}d)$, where $d$ is the dimension of $\mathbf{z}$ and $\mathbf{c}$. Typically, for a $5$-second sequence, $K$ ranges from $10$ to $20$, while $T$ is around 120. Compared to methods that assemble every frame individually with complexity $O(T^2d)$, our approach uses $\frac{T^2}{K^2} \ll T^2$ and $d \ll N$, further demonstrating the efficiency of our 4D representation.

Compared to single global code methods, $\mathcal{M}$ allows the diffusion model to directly operate on the latent of a specific frame, avoiding drift from single global code and resulting in high-fidelity volumetric 4D sequences. Moreover, this also allows us to apply attention mechanisms to the matrix for shape-to-color supervision, as well as conditional generation. This preserves advantage of our coherent 3D shape and color modeling, namely to allow the shape to supervise the color while also avoiding a fixed combination of shape and color.

Our matrixized 4D representations indicate that, although the initial number of keyframes for one 4D sequence is fixed, variable-length 4D sequences can be obtained from two or more 4D sequences through recovering the unseen frames via Eq.~\ref{eq:linear_intepl} and ~\ref{eq:map_intepl}. Once the latent representation is generated, our method allows for flexible decoding, enabling the reconstruction of variable-length volumetric 4D sequences by interpolating latent keyframes in Fig.~\ref{fig:model}.

\label{sec:diffusion}
\subsection{Spatio-Temporal Diffusion for Conditional 4D Volumetric Generation}
With the proposed latent 4D sequence representation, we can interpolate to achieve a volumetric 4D sequence with variable frame density. We then use a multi-condition diffusion model to generate the latent 4D sequence representation under the given conditions.

\noindent\textbf{Condition Fusion.}
Our diffusion model generates a 4D latent representation conditioned by image-text input (see Fig.~\ref{fig:model}). To maintain the consistency of animation but also enable the different combinations of motion, we apply a description template $d_i$ embedded with frame index $i$ to condition
coherent sequences(e.g. the third frame of some posed objects). Additionally, to maintain visual consistency, we replicate the image features, using the same image condition $I$ for generation across each frame:
\begin{equation}
    f_i^T = \tau_\beta(d_i \circ t_i) \qquad f_i^I = \tau_\theta(I),
\end{equation}
where $\tau_\beta$ and $\tau_\theta$ are the CLIP~\cite{CLIP} text and image encoders, $t_i$ is the text condition of the $i$-th frame, and $\circ$ means concatenation.

\begin{figure}[t]
    \centering
    \includegraphics[width=0.7\linewidth]{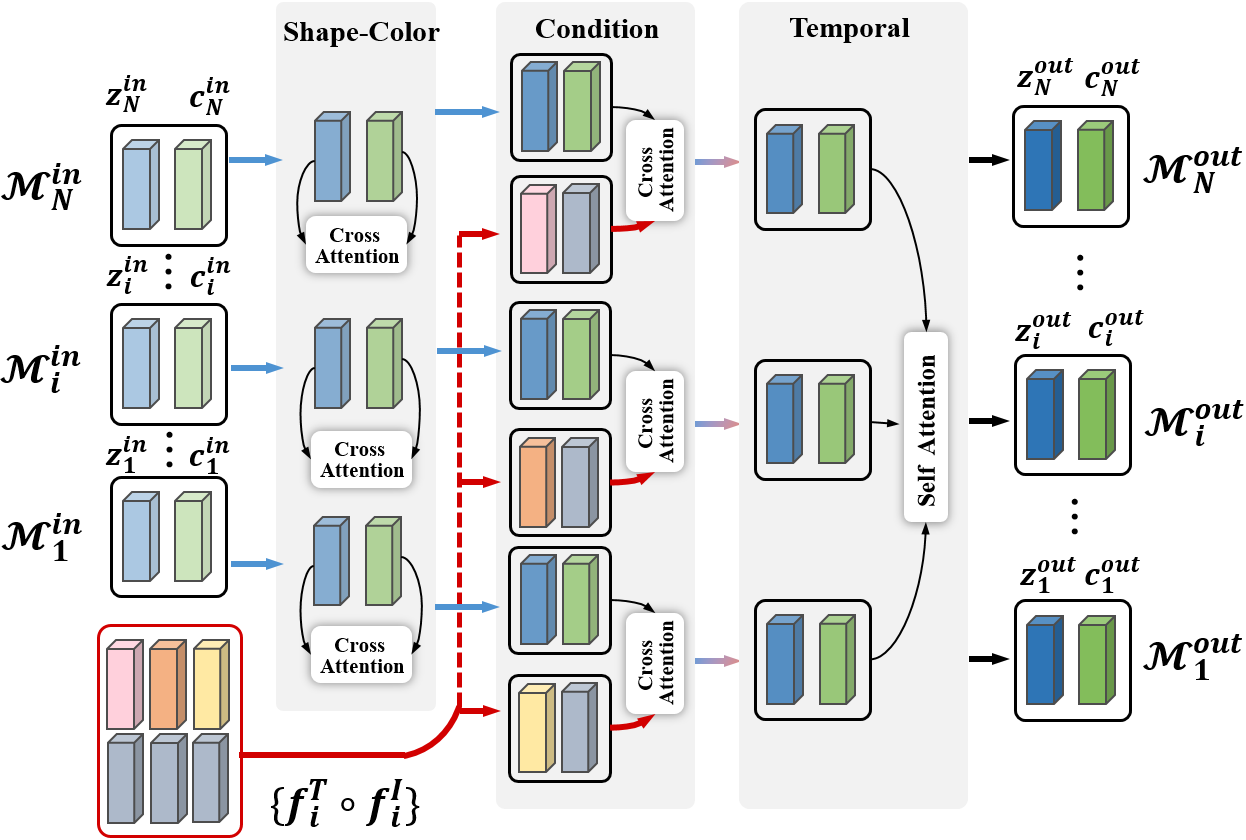}
    \vspace{-4mm}
    \caption{Hierarchical Conditional Spatio-Temporal Attention (HCSTA) block, repeatedly applying to get the final denoised $\mathcal{M}$. Within each layer, different colored latents represent the dynamics of distinct local regions, while the same colored latents represent the dynamics of a local region at different time steps.}
    \label{fig:attention-show}
    \vspace{-5mm}
\end{figure}

\noindent\textbf{Hierarchical Conditional Spatio-Temporal Attention (HCSTA).}
In the diffusion process, we aim to (1) color latent representation is conditioned on shape latent representation, (2) the model should capture the evolution of latent representations between frames, (3) conditions can correctly guide the generation of $\mathcal{M}$. Our latent 4D sequence representation avoids brute-force attention across spatial and temporal domains. Leveraging this, we designed the HCSTA to meet these goals, as shown in Fig.~\ref{fig:attention-show}. Each HCSTA block has three attention layers: shape-color cross-attention layer, conditioned cross-attention layer, and temporal self-attention layer.
The shape-color cross-attention layer enables the color latent codes $\{\mathbf{c}_i^\mathrm{in} \in \mathbb{R}^{d_2}\}$ to be supervised by the shape latent codes $\{\mathbf{z}_i^\mathrm{in} \in \mathbb{R}^{d_1}\}$:
\begin{equation}
    \mathcal{A}_\mathrm{shape-color} = \mathrm{CrossAttn}(\{\mathbf{c}_i^\mathrm{in} \in \mathbb{R}^{d_2}\}, \{\mathbf{z}_i^\mathrm{in} \in \mathbb{R}^{d_1}\}).
\end{equation}
Then, the conditioned cross-attention layer injects the conditional information $f_i^T \circ f_i^I$ into the entire 4D representation $\mathcal{M}^\mathrm{in}$:
\begin{equation}
    \mathcal{A}_\mathrm{cond} = \mathrm{CrossAttn}(\mathcal{M}^\mathrm{in}, \{f_n^T \circ f_n^I\}).
\end{equation}
Finally, to enhance spatio-temporal consistency, the temporal self-attention layer applies self-attention across different frames along the temporal dimension of $\mathcal{M}^{in}$:
\begin{equation}
    \mathcal{A}_\mathrm{time} = \mathrm{SelfAttn}(\mathcal{M}^\mathrm{in}).
\end{equation}

During the denoising phase, we implement classifier-free guidance for the image-text conditional generation of the 4D representation $\mathcal{M}^t$ at denoising step $t$. By treating the latent 4D sequence representation as a cohesive entity, we apply a uniform noise reduction strategy throughout, ensuring the preservation of local pattern consistency. 
The simplified optimization objective follows the approach outlined in ~\cite{rombach2022high}:
\begin{equation}
    \begin{aligned}
        \mathcal{L}_{LDM} = \mathop\mathbb{E}_{\mathcal{M}, f, \epsilon, t} \left[ {\epsilon - \epsilon_\theta(\mathcal{M}^t, t, F\{D\circ \tau_i(f_i)\}) }^2 \right],
    \end{aligned}
\end{equation}
where $D$ is a dropout operation enabling classifier-free guidance, $\tau_i(f_i)$ is the $i$-th modality feature encoded by the task-specific encoder $\tau_i$, $\epsilon_\theta$ is the denoiser equipped with our HCSTA module, and $F$ refers to a simple concatenation. This approach not only ensures uniformity in the denoising process but also significantly reduces computational overhead.

\section{Experiments}

\noindent\textbf{Data Preparation.} For 3D geometry evaluation, we use car, chair, and airplane categories from ShapeNet \cite{chang2015shapenet}. For colored 3D objects evaluation, we use T-pose objects from 3DBiCar \cite{luo2023rabit}. We use 16-frame animal sequences from DeformingThings4D (DT4D) \cite{li20214dcomplete} and 20-frame colored cartoon sequences from 3DBiCar (sample $30$k sequences via latent space interpolation) for both unconditional and conditional generation. DeformingThings4D~\cite{li20214dcomplete} is a synthetic dataset containing $1,972$ animation sequences spanning 31 categories of humanoids and animals. However, the lack of color information and image annotations in DeformingThings4D makes it challenging to use for conditional 4D sequence generation. The 3DBiCar~\cite{luo2023rabit} dataset provides a script for sampling high-quality, topologically consistent 3D textured models of biped cartoon characters, manually crafted by professional artists, covering diverse identities, shapes, and textures across $15$ species and $4$ image styles. We also use $110$k subset and $323$k of Objaverse~\cite{deitke2023objaverse} for large scale conditional 4D generation. Objaverse is a large-scale 3D dataset containing both static and dynamic objects with more complex materials and geometric details, making it suitable for evaluating model performance on large-scale data. We select high-quality 4D sequences from the $110$k and $323$k subsets to train our method.

Both the 3DBiCar and Objaverse datasets provide image descriptions with masks relatively easily, but most of the data lack textual annotations. We employ Large Language Models (LLMs) for initial image descriptions, manually refined for accuracy in shape, color, and pose, with plans to release these captions post-acceptance. For conditioning, masked images were derived from provided masks and replicated across frames, while text conditions incorporated frame-order prefixes (e.g., "$n$-th frame") and were encoded with CLIP alongside image conditions. Training data included T-pose to target-pose transitions for model optimization.

    


\noindent\textbf{Evaluation Metrics.} We follow prior works~\cite{erkocc2023hyperdiffusion, cheng2023sdfusion} in evaluating Minimum Matching Distance (MMD), Coverage (COV), and 1-Nearest-Neighbor Accuracy (1-NNA):
\begin{equation}
    \text{MMD}(S_g, S_r) = \frac{1}{\vert S_r \vert} \sum_{Y \in S_r} \min_{X \in S_g} D(X, Y)
\end{equation}
\begin{equation}
    \text{COV}(S_g, S_r) = \frac{\vert \{ \arg\min\limits_{Y \in S_r} D(X, Y) \vert X \in S_g \} \vert}{\vert S_r \vert}
\end{equation}
\begin{small}
\begin{equation}
    \begin{aligned}
    \text{1-NNA}(S_g, S_r) =  \frac{\sum_{X \in S_g} \Indicator[N_X \in S_g] + \sum_{Y \in S_r} \Indicator[N_Y \in S_r] }{\vert S_g \vert + \vert S_r \vert},
    \end{aligned}
\end{equation}
\end{small}
where $D(X, K)$ is some kind of distance measure, and in the 1-NNA metric $N_X$ is a point cloud that is closest to $X$ in both generated and reference dataset: 
\begin{equation}
    N_X = \arg\min\limits_{K \in S_r \cup S_g} D(X, K)
\end{equation}
where $S_r$ is the reconstructed 4D sequence, $S_g$ is the ground-truth 4D sequence. We also use a Chamfer Distance (CD) distance measure $D(X,Y)$ for computing these metrics in 3D, and report CD values multiplied by a constant $10^2$. 
To evaluate 4D shapes, we extend to the temporal dimension for $T$ frames:
\begin{equation}
    D(X, Y) = \frac{1}{T}\sum_{t=0}^{T - 1} CD(X[t], Y[t]).
\end{equation}

For MMD and CD, lower is better; for COV, higher is better; for 1-NNA, 50\% is the optimal. For \textit{4D sequence evaluation}, we average 3D metrics across frames. To assess \textit{inter-frame consistency}, we compute CD-VAR and COV-VAR by calculating Chamfer Distance and COV variance across frames. For \textit{color evaluation}, PSNR and SSIM are computed by sampling 16,384 surface points multiple times. We also use CLIP score (CLIP-S). All videos are rendered from Blender.

\noindent\textbf{Implementation Details.} The $\sigma$ is computed as $0.3\%$ of the original 4D sequence reconstruction loss. The loss weights $\lambda_1$ and $\lambda_2$ are set to $1$. For the shape auto-decoder $D_\theta$, we employ an 8-layer MLP to decode the SDF values, using a resolution of $64 \times 64 \times 64$ to construct a 6-layer 3D U-Net $U_\phi$ on DT4D and $128 \times 128 \times 128$ on 3DBiCar, ShapeNet, and Objaverse. For the color auto-decoder $C_\theta$, we use an 8-layer MLP. For denoising network, we use an 8-layer UNet augmented with HCSTA Layer.

\subsection{Ablation Studies}
\noindent\textbf{Dimension of latents and loss weights.} We provide an overview of the hyperparameter of our models in Tab.~\ref{tab:hyperparameters}, including the size of the latent 4D representation $\mathcal{M}$ and loss weights. We evaluated the metrics and finally selecting the combination $\{\mathcal{M}\in\mathbb{R}^{N\times(256+256)}, ~\lambda_1=1.0, ~\lambda_2=1.0\}$. Because balancing the L1 and SSIM losses aids the model in learning both the primary colors and the finer color details in a more balanced manner. Moreover, experimentation with dimensions larger than the current setting (256+256) was conducted. However, it is observed that when the scale of $\mathcal{M}$ exceeds 256 (as indicated in row 3 of Tab.~\ref{tab:hyperparameters}), the data distribution becomes too sparse to support the diffusion generation for volumetric 4D sequence effectively. 

\begin{figure}[t]
    \centering
    \begin{minipage}[c]{0.48\linewidth} 
        \centering
        \scriptsize
        \setlength\tabcolsep{12pt}
        \begin{tabular}{ccccc | c}
        \hline \hline
        $\mathcal{M}$                       & $\lambda_1$              & $\lambda_2$                   & CLIP-S        \\ \midrule
        $\mathbb{R}^{N\times(256+256)}$     & 1.0                      & 1.0                           &0.911          \\ 
        $\mathbb{R}^{N\times(128+128)}$     & 1.0                      & 1.0                           & 0.733         \\ 
        $\mathbb{R}^{N\times(512+512)}$     & 1.0                      & 1.0                           & 0.909         \\ 
        $\mathbb{R}^{N\times(256+256)}$     & 0.5                      & 1.0                           & 0.787         \\ 
        $\mathbb{R}^{N\times(256+256)}$     & 1.0                      & 0.5                           & 0.691         \\ \hline \hline
        \end{tabular}
        \vspace{-2mm}
        \captionof{table}{Ablation of hyperparameter.}
        \label{tab:hyperparameters}

        \setlength\tabcolsep{2pt}
        \begin{tabular}{c c | ccccc}
        \hline \hline
        Category & $U_\phi$ & MMD & CD-VAR & COV(\%) & COV-VAR & 1-NNA(\%) \\ \midrule
                 & $\times$     & 0.012          & 0.007          & 69  & 2.791  & 36 \\
        \multirow{-2}{*}{3DBiCar} & $\checkmark$ & \textbf{0.009} & \textbf{0.004} & \textbf{73} & \textbf{0.834} & \textbf{41} \\ \midrule
                 & $\times$     & 0.037          & 0.003          & 69  & 3.179  & 36 \\
        \multirow{-2}{*}{DT4D}    & $\checkmark$ & \textbf{0.030} & \textbf{0.001} & \textbf{76} & \textbf{2.91}  & \textbf{43} \\ \hline \hline
        \end{tabular}
        \vspace{-2mm}
        \captionof{table}{Ablation study of $U_\phi$.}
        \label{tab:abl-4d-refine-comp}
    \end{minipage}
    \hfill
    \begin{minipage}[c]{0.48\linewidth} 
        \centering
        \scriptsize
        \setlength\tabcolsep{4pt}
        \begin{tabular}{cc | cccc}
        \hline \hline
        Category & Method & MMD & COV(\%) & 1-NNA(\%) & CLIP-S \\ \midrule
                 & DeepSDF  & 0.264 & 63.11 & 71.11 & - \\
                 & SDFusion & 0.088 & 29.8  & 83.71 & 0.823 \\
        \multirow{-3}{*}{Airplane} & Ours & \textbf{0.032} & \textbf{65.22} & \textbf{61.29} & \textbf{0.897} \\ \midrule
                 & DeepSDF  & 0.240 & 64.53 & 68.85 & - \\
                 & SDFusion & 0.599 & 20.59 & 54.85 & 0.842 \\
        \multirow{-3}{*}{Car} & Ours & \textbf{0.201} & \textbf{66.45} & \textbf{54.45} & \textbf{0.866} \\ \midrule
                 & DeepSDF  & 0.368 & 90.42 & 18.93 & - \\
                 & SDFusion & 0.184 & 50.83 & 84.85 & 0.869 \\
        \multirow{-3}{*}{Chair} & Ours & \textbf{0.04} & \textbf{95.56} & \textbf{33.91} & \textbf{0.885} \\ \midrule
                 & DeepSDF  & 0.112 & 71.98 & 38.46 & - \\
                 & SDFusion & 0.076 & 73.19 & 40.03 & 0.812 \\
        \multirow{-3}{*}{3DBiCar} & Ours & \textbf{0.067} & \textbf{74.63} & \textbf{41.92} & \textbf{0.911} \\ \hline \hline
        \end{tabular}
        \vspace{-2mm}
        \captionof{table}{Quantitative comparison of Single Shape.}
        \label{tab:single-shape}
    \end{minipage}
\end{figure}

Finally, analysis of the remaining parameter settings (rows 4-5) reveals that an unbalanced weighting of loss terms significantly leads to performance degradation. 

\noindent\textbf{Refinement Module $U_\phi$.} First, we compare 3D shape reconstruction quality among SDFusion~\cite{cheng2023sdfusion}, DeepSDF~\cite{park2019deepsdf}, and our method. SDFusion uses a 3D latent diffusion model with a VQVAE decoder for shape generation and a 2D diffusion model for texture rendering. DeepSDF decodes SDF values without color using an MLP decoder. We evaluate geometry accuracy on ShapeNet (car, chair, airplane) and 3DBiCar (T-pose) datasets. Tab.~\ref{tab:single-shape} shows our method outperforms SDFusion and DeepSDF in geometry generation and surpasses SDFusion in color generation, confirming that the refinement U-Net $U_\phi$ improves the quality of coarse SDF with a 3D U-Net, making our coherent 3D shape and color modeling learn more shape details. Then we ablate $U_\phi$ in 4D sequence generation, as demonstrated in Tab.~\ref{tab:abl-4d-refine-comp}, where $U_\phi$ improves all metrics because $U_\phi$ enhances the geometric performance of all frames within the 4D sequence, thereby indirectly promoting consistency across frames. 

\noindent\textbf{HCSTA module. } Tab.~\ref{tab:abla-hcsta} shows ablation studies on attention modulation, order, and unification. We first evaluate attention for shape-color (S), conditioned (C), and temporal (T) dimensions. Tab.~\ref{tab:abla-hcsta} (rows 1-3) shows shape-color cross-attention contributes most. In rows 4-7, we permute the S, C, T orders, the S-C-T order performing best. Because our color field is geometry-aware, where accurate shape modeling facilitates precise subsequent generation of both color and motion. Finally, replacing the proposed hierarchical attention with a unified strategy (shared attention) leads to a significant drop (rows 8-9), our method outperforms it. Because our decoupled latent and attention design enables the model to explicitly learn accurate shape, color, and spatiotemporal relationships, leading to significant performance gains.
\begin{table}[t]
    \normalsize
    \centering
    \resizebox{0.85\textwidth}{!}{
	\begin{tabular}{c c c c | ccccc}
		\hline \hline
		Type & S &C &T                    & MMD            &CD-VAR         & COV(\%)     &COV-VAR           & 1-NNA(\%) \\
		\midrule 
        \multirow{3}{*}{Module} & $\times$ & $\checkmark$ &$\checkmark$ & 0.014 & 0.018 & 61 & 2.877 & 33 \\
        & $\checkmark$ & $\times$ &$\checkmark$ & 0.010 & 0.015 & 68 & 2.165 & 39   \\
        & $\checkmark$ & $\checkmark$ &$\times$ & 0.011 & 0.011 & 71 & 2.663 & 38   \\
        \midrule
         \multirow{3}{*}{Order} & \multicolumn{3}{c|}{S-C-T} & \textbf{0.009} & \textbf{0.007} & \textbf{73} & \textbf{1.834} & \textbf{41} \\
         & \multicolumn{3}{c|}{S-T-C}& 0.010 & 0.011 & 71 & 2.165 & 39  \\
         & \multicolumn{3}{c|}{C-T-S}& 0.012 & 0.015 & 68 & 2.663 & 38  \\
         & \multicolumn{3}{c|}{C-S-T}& 0.011 & 0.014 & 69 & 2.731 & 35  \\ \midrule
         \multirow{2}{*}{Uni-Stra.}
         & \multicolumn{3}{c|}{$\checkmark$}  & 0.013 & 0.018 & 67 & 3.002 & 34 \\
         & \multicolumn{3}{c|}{$\times$}& \textbf{0.009} & \textbf{0.007} & \textbf{73} & \textbf{1.834} & \textbf{41} \\ \hline \hline
	\end{tabular}}
    \vspace{-3mm}
     \caption{\label{tab:abla-hcsta} Ablation on attentions modulation, orders, and unifying.}    
 \vspace{-3mm}
\end{table}

\noindent\textbf{Coherent 3D Shape and Color Modeling.} 
We evaluate the impact of L1 loss, remeshing (R), structural similarity loss (SSIM), and a shape-agnostic approach (F) that assigns colors from a uniform distribution per object. This strategy reverses the marching cubes algorithm to sample color values at uniformly distributed points within the query space, such as a $64\times64\times64$ resolution grid:
\begin{gather}
    \alpha = \frac{v_2-v_{iso}}{v_2-v_1} \\
    \mathbf{C}^i = \alpha\mathbf{C}^i_1+(1-\alpha)\mathbf{C}^i_2
\end{gather}
where $\alpha$ is the linear interpolation weight, $v_1$ and $v_2$ are the SDF values of  $p_1$ and $p_2$, $v_{iso}$ is the isolated value used for reconstructing the isosurface of an object, $\mathbf{C}^i_1$ and $\mathbf{C}^i_2$ are color of $p_1$ and $p_2$. We fix $\mathbf{C}^i_2$ to get $\mathbf{C}^i_1$. The colors of other query coordinates that do not contribute to the reconstruction vertices are obtained through linear interpolation. We apply the same training strategy to the geometric coherent color representation learning pipeline.

\begin{figure}[t]
    \centering
    \begin{minipage}[c]{0.48\linewidth}
        \centering
        \includegraphics[width=0.85\linewidth]{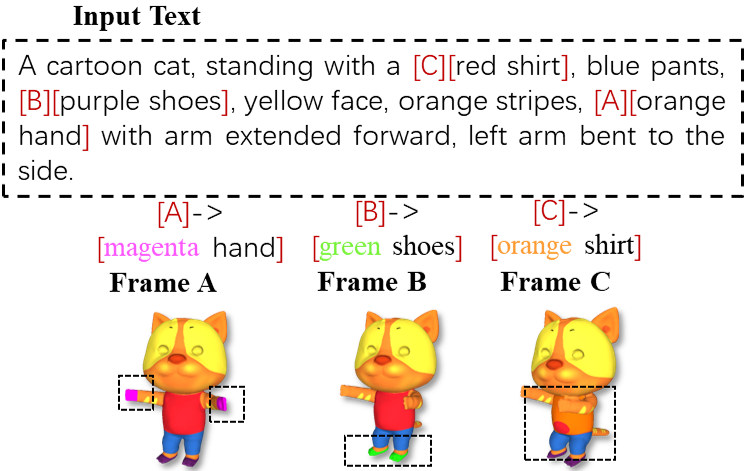}
        \vspace{-1.5mm}
        \caption{Color editing without shape alteration.}
        \label{fig:suppl-decouple}
    \end{minipage}
    \begin{minipage}[c]{0.48\linewidth}
        \centering
        \footnotesize
        \setlength\tabcolsep{8pt}
        \begin{tabular}{c c c | c || cc}
            \hline \hline
            L1 & R & SSIM & F & PSNR & SSIM \\ \midrule
            $\checkmark$ & $\checkmark$ & $\times$     & $\times$     & 29.15          & 0.79          \\
            $\checkmark$ & $\times$     & $\checkmark$ & $\times$     & 26.37          & 0.72          \\
            $\times$     & $\checkmark$ & $\checkmark$ & $\times$     & 28.76          & 0.75          \\
            $\checkmark$ & $\checkmark$ & $\checkmark$ & $\checkmark$ & 9.64           & 0.64          \\ \midrule
            $\checkmark$ & $\checkmark$ & $\checkmark$ & $\times$     & \textbf{34.51} & \textbf{0.85} \\ \hline \hline
        \end{tabular}
        \vspace{-2mm}
        \captionof{table}{Quantitative results of color ablation study.}
        \label{tab:abl-l1-ssim-comp}
    \end{minipage}
\end{figure}

As evidenced in Tab.~\ref{tab:abl-l1-ssim-comp}, integrating L1, R, and SSIM optimizes color fidelity: L1 ensures  the overall color accuracy, SSIM improves local structural coherence, and R enhances vertex density for refined color detail. Our coherent 3D shape and color modeling constructs a continuous color field adaptive to varied geometries, ensuring shape-constrained color continuity and boosting accuracy. In contrast, shape-agnostic color learning severely weakens the correlation between color and geometry, leading to a significant drop in color fitting quality.

\noindent\textbf{Decoupled Color and Shape.} 
We validate the efficacy of our decoupled shape and color framework by assessing fitting quality for combined and decoupled latents in Tab.~\ref{tab:fitting-quality}. Conditional 4D generation quality and shape-color consistency are further evaluated through a user study in Tabs.~\ref{tab:decouple-user-study}, where participants compared our method to baselines across three metrics, selecting the superior approach per vote. Fig.~\ref{fig:suppl-decouple} demonstrates that the decoupled representation allows color editing without altering shape. Results in Tabs.~\ref{tab:decouple-user-study} consistently show that ours improves generation quality and enables independent color modifications. The explicit decoupling of shape and color allows for stable and independent optimization of structural and appearance components, thereby significantly enhancing intra-frame consistency by preventing color drifting and boosting inter-frame coherence by enabling smooth and geometry-aware temporal transitions.

\begin{table}[t]
    \centering
    \scriptsize
    \setlength\tabcolsep{5pt}
    \begin{tabular}{c}
        \begin{minipage}{0.48\linewidth}
            \centering
            \begin{tabular}{c|ccccc}
                \hline \hline
                Method & MMD & COV(\%) & PSNR & SSIM & CLIP-S\\ \midrule
                Combined & 0.021 & 57 & 29.13 & 0.72 & 0.801\\
                \textbf{Decoupled} & \textbf{0.009} & \textbf{73} & \textbf{34.51} & \textbf{0.85} & \textbf{0.911} \\ \hline \hline
            \end{tabular}
            \captionof{table}{Ablation of combined and decoupled representations in fitting.}
            \label{tab:fitting-quality}
        \end{minipage}
        \hspace{4mm}
        \begin{minipage}{0.48\linewidth}
            \centering
            \begin{tabular}{c|cc}
                \hline \hline
                Metric & Combined & \textbf{Decoupled} \\ \midrule
                Color and Shape Consistency & 7.33\% & \textbf{92.67\%} \\
                Interframe Consistency & 6.31\% & \textbf{93.69\%} \\ \hline \hline
            \end{tabular}
            \captionof{table}{User study voting rates on color, shape, and interframe consistency.}
            \label{tab:decouple-user-study}
        \end{minipage}
    \end{tabular}
\end{table}

\vspace{-2mm}
\subsection{Unconditional 4D Sequence Shape Generation}
We first demonstrate the capability of our model to generate diverse objects and motions in unconditional 4D generation, then we showcase its ability to produce coherent sequences through latent matrixized 4D representation interpolation. We conduct evaluation on 3DBiCar and DeformingThings4D datasets, focusing on shape and temporal consistency. Tab.~\ref{tab:4d-comp} presents quantitative comparisons of 4D sequence generation, demonstrating that our model consistently outperforms the voxel baseline and HyperDiffusion~\cite{erkocc2023hyperdiffusion} (Hyper.) across all evaluated metrics. Because the voxel baseline and HyperDiffusion face challenges related to voxel resolution, which limits geometric precision, and high-dimensional sparsity in the weight space, which hindering HyperDiffusion from reconstructing high-quality 4D sequences, our method improves unconditional 4D generation by employing a matrixized 4D representation, which reduces the learning difficulty of denoiser during the diffusion process.

\begin{figure}[t]
    \centering
    \begin{minipage}[c]{0.52\linewidth}
        \centering
        \includegraphics[width=0.95\linewidth]{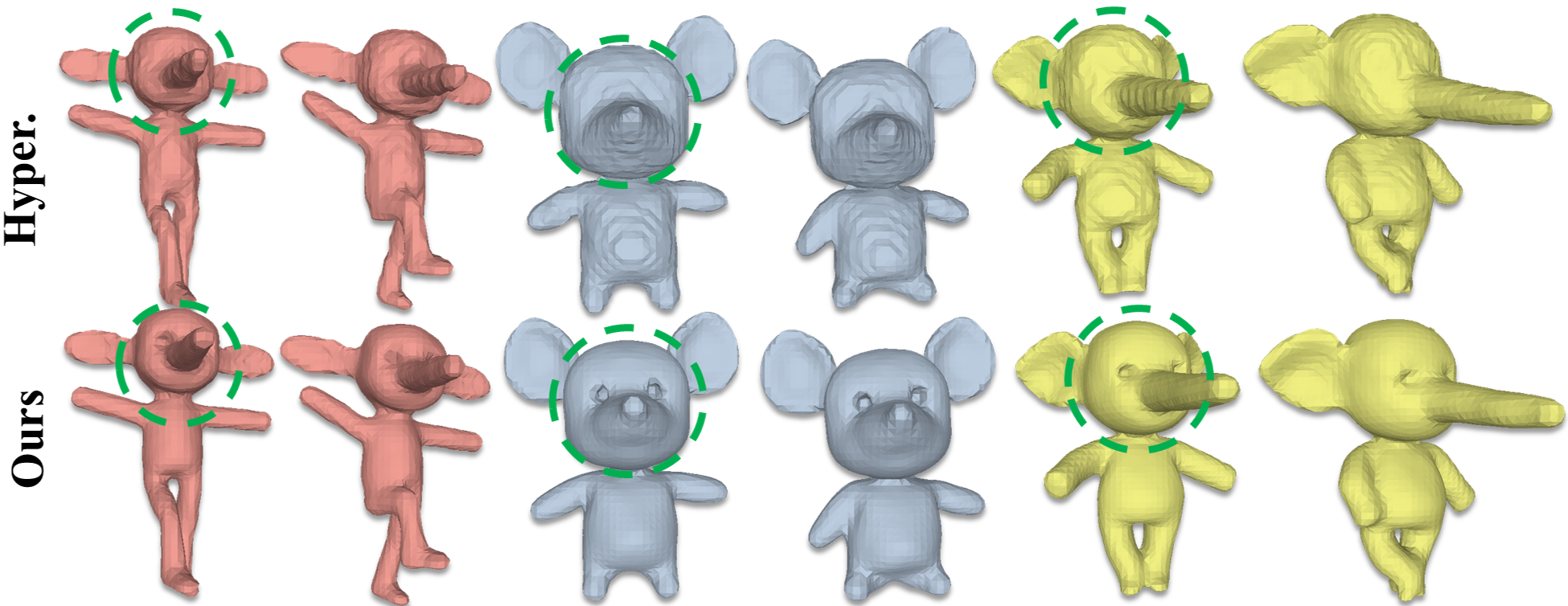}
        \vspace{-3mm}
        \caption{Unconditional 4D Generation Comparison.}
        \label{fig:unconditonal-4d-1}
    \end{minipage}
    \begin{minipage}[c]{0.45\linewidth}
        \centering
        \includegraphics[width=0.9\linewidth]{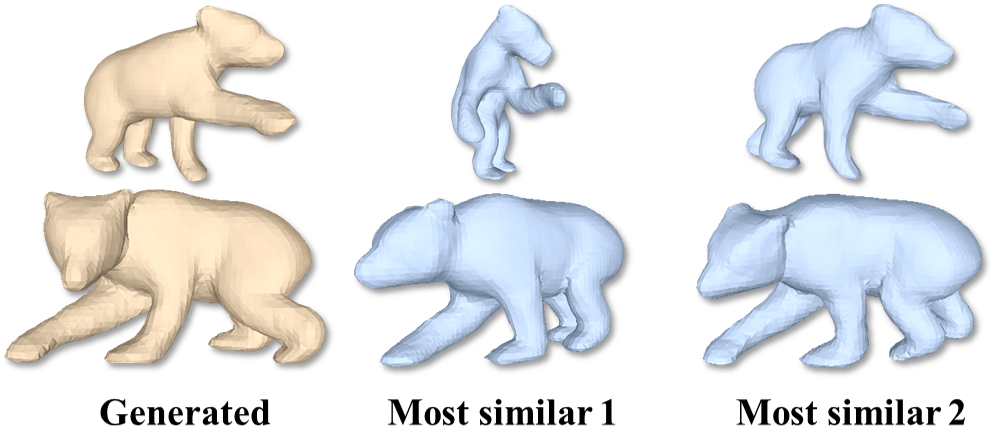}
        \vspace{-0.5mm}
        \caption{Nearest sampling via CLIP score.}
        \label{fig:suppl-near}
    \end{minipage}
    \vspace{-2mm}
\end{figure}

Fig.~\ref{fig:unconditonal-4d-1} provides a qualitative comparison of 4D sequences generated by HyperDiffusion and our method, illustrating that our model produces smoother surfaces with more detailed features, such as facial details. This highlights our superior ability to generate 4D sequences with refined geometric details. We also show the most similar objects in DT4D dataset to unconditional generation results in Fig.\ref{fig:suppl-near}, demonstrating the ability of our method to generate diverse 4D sequences.

\begin{table}[t]
    \centering
    \scriptsize
    \centering
    \scriptsize
    \small
    \setlength\tabcolsep{6pt}
    \begin{tabular}{cc| ccccc}
    \hline \hline
    Category                            & Method                    & MMD            &CD-VAR         & COV(\%)     &COV-VAR           & 1-NNA(\%) \\ \midrule
                                        & Voxel                     & 0.293          &0.035          & 30          &1.857             & 91   \\
                                        & Hyper.~\cite{erkocc2023hyperdiffusion}           & 0.044          &0.011          & 52          &2.138             & 69.3 \\
    \multirow{-3}{*}{3DBiCar}           & Ours                      & \textbf{0.009} &\textbf{0.007}          & \textbf{73} &\textbf{1.834}             & \textbf{41} \\ \midrule
                                        & Voxel                     & 0.219          &0.044          & 35          &1.673             & 85   \\
                                        & Hyper.~\cite{erkocc2023hyperdiffusion}            & 0.155          &0.013          & 45          &1.735             & 62   \\
    \multirow{-3}{*}{DT4D} & Ours                      &\textbf{0.030}  &\textbf{0.001} & \textbf{76} &\textbf{2.91}              &\textbf{43}  \\ \hline \hline
    \end{tabular}
    \vspace{-3mm}
    \caption{
        Quantitative comparison of 4D Sequence Generation.
    }
    \label{tab:4d-comp}
\end{table}

\begin{figure}[t]
    \centering
    \includegraphics[width=0.8\linewidth]{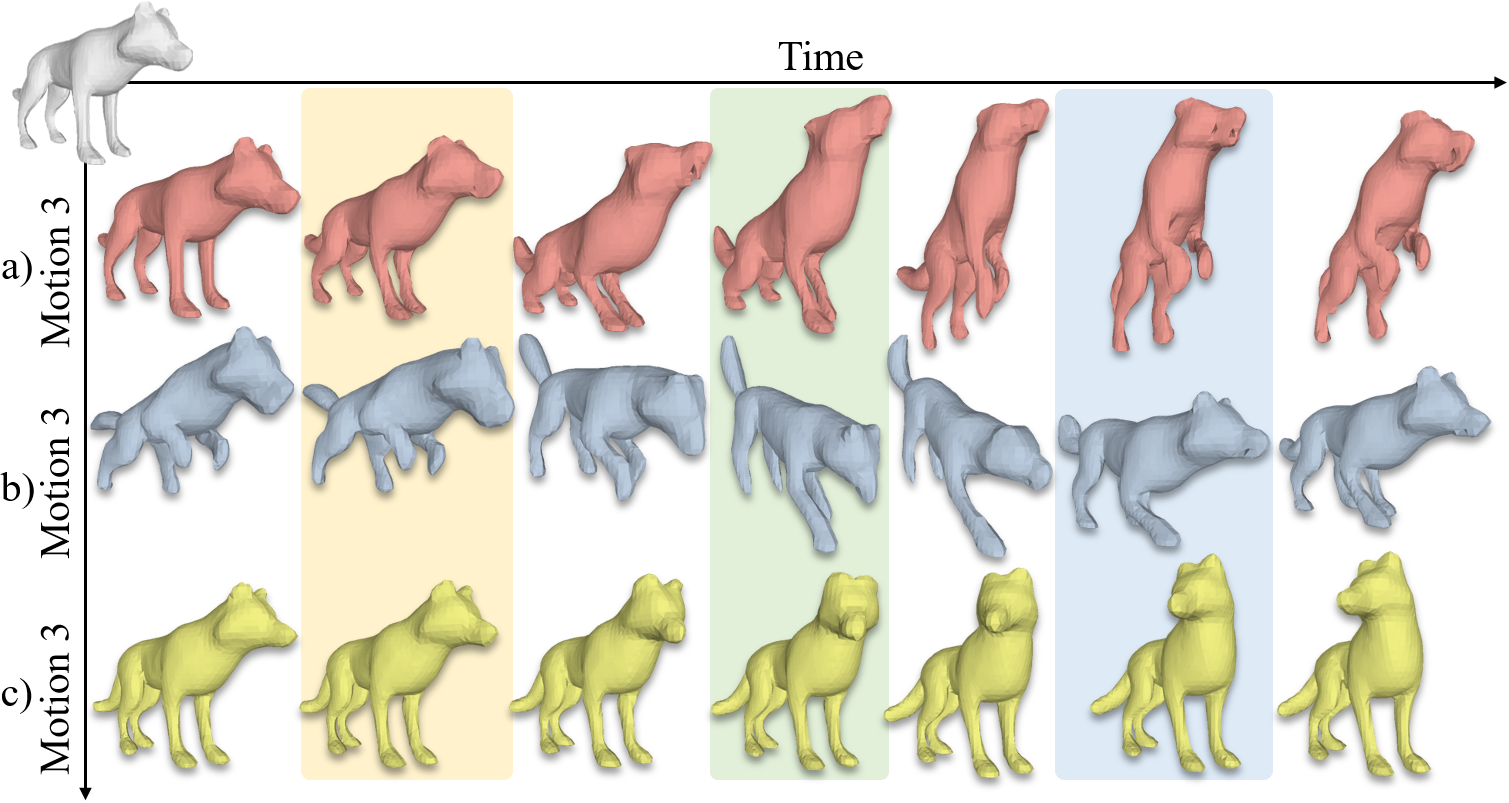}
    \vspace{-5mm}
    \caption{Diverse 4D motion sequences generated and interpolated from the model trained on DeformingThings4D.
    }
    \label{fig:interpolation}
    \vspace{-4mm}
\end{figure}

Fig.~\ref{fig:interpolation} explores motion diversity and interpolation, with each row depicting a different action performed by the same object. This demonstrates our capability to generate varied motions for the same object with high quality, highlighting both its versatility and its detail-rich output. Additionally, the 2nd, 4th, and 6th columns display objects decoded from interpolated $\mathcal{M}$. A comparison of these columns with their adjacent ones reveals that the interpolated results exhibit strong temporal consistency, indicating that our latent matrixized 4D representation supports the generation of sequences with variable-length frame density through interpolation. Because our matrixized 4D representation forms a continuous and meaningful latent space after sufficient training, enabling the construction of variable-length 4D sequences through interpolation.

\vspace{-2mm}
\subsection{Conditonal 4D Sequence Generation}

\begin{figure}[t]
    \centering
    \includegraphics[width=0.8\linewidth]{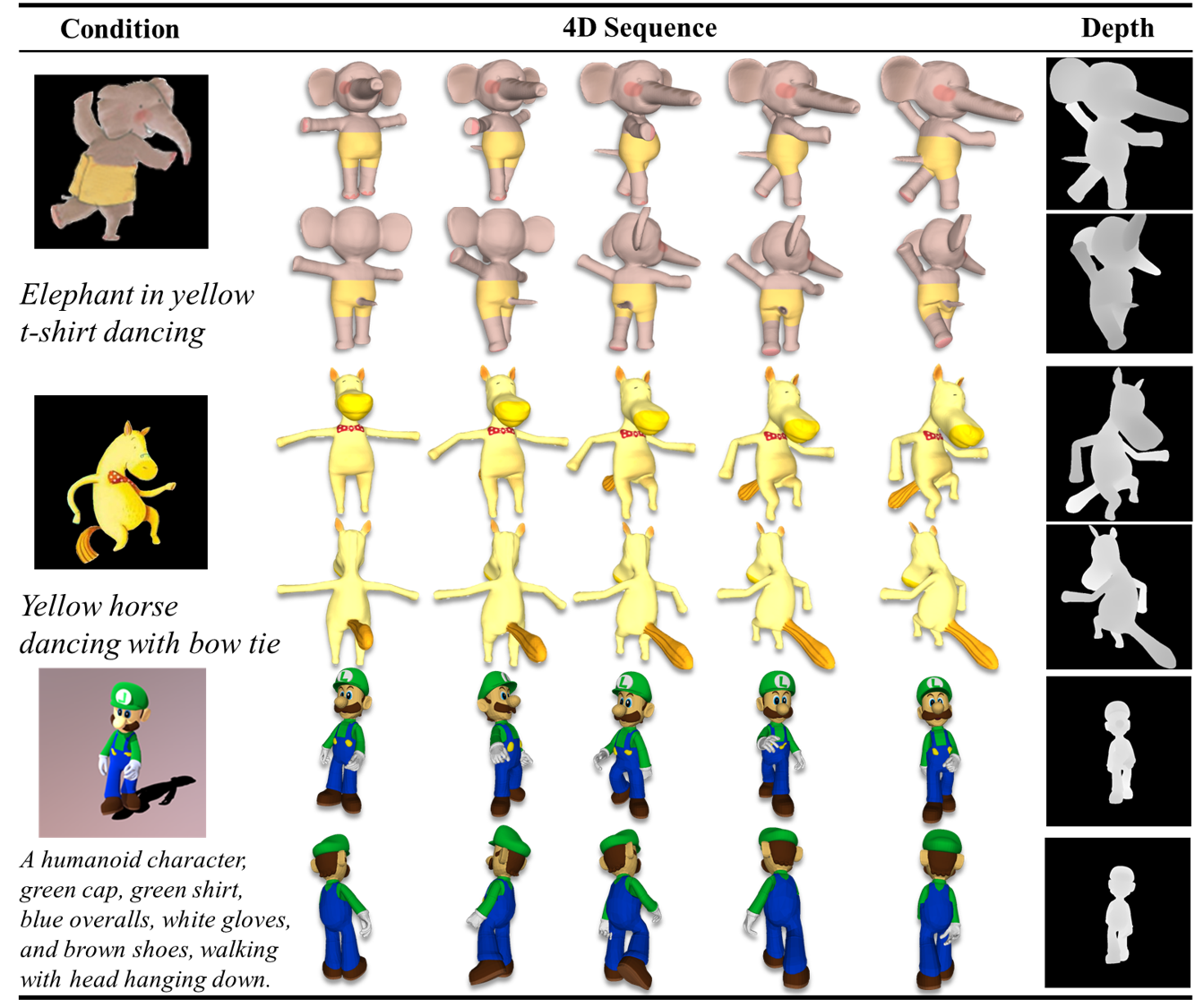}
    \vspace{-2mm}
    \caption{Conditional 4D Sequence Generation visulizations on 3DBiCar and Objaverse.}
    \label{fig:conditonal-4d}
    \vspace{-2mm}
\end{figure}

\begin{figure*}[t]
    \centering
    \includegraphics[width=0.85\linewidth]{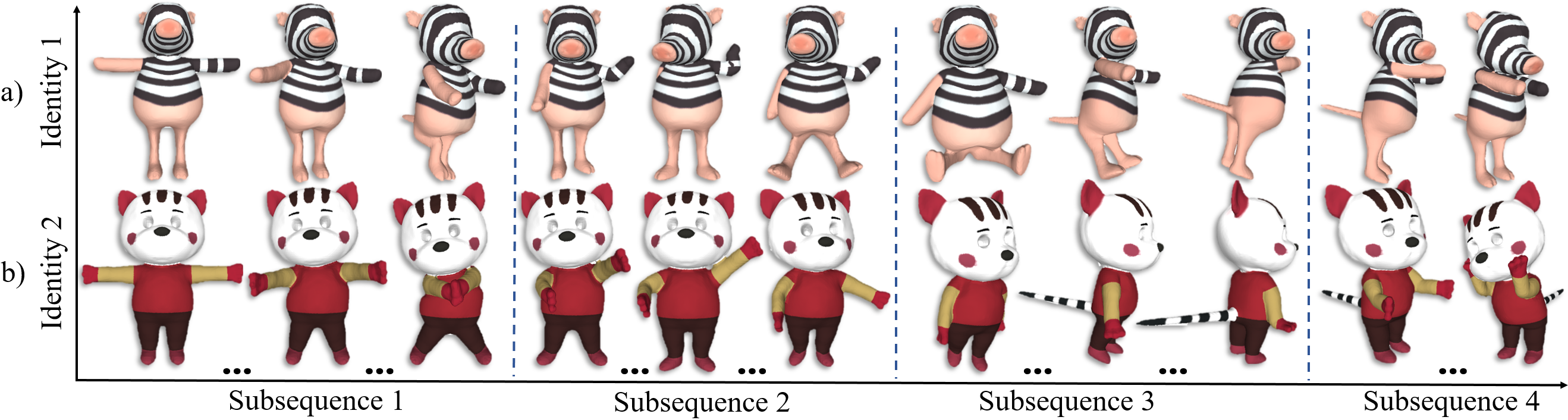}
    \vspace{-3mm}
    \caption{Long sequence 4D generation via latent frames interpolation.}
    \label{fig:long-seq}
    \vspace{-3mm}
\end{figure*}

\begin{figure}[t]
    \centering
    \includegraphics[width=0.8\linewidth]{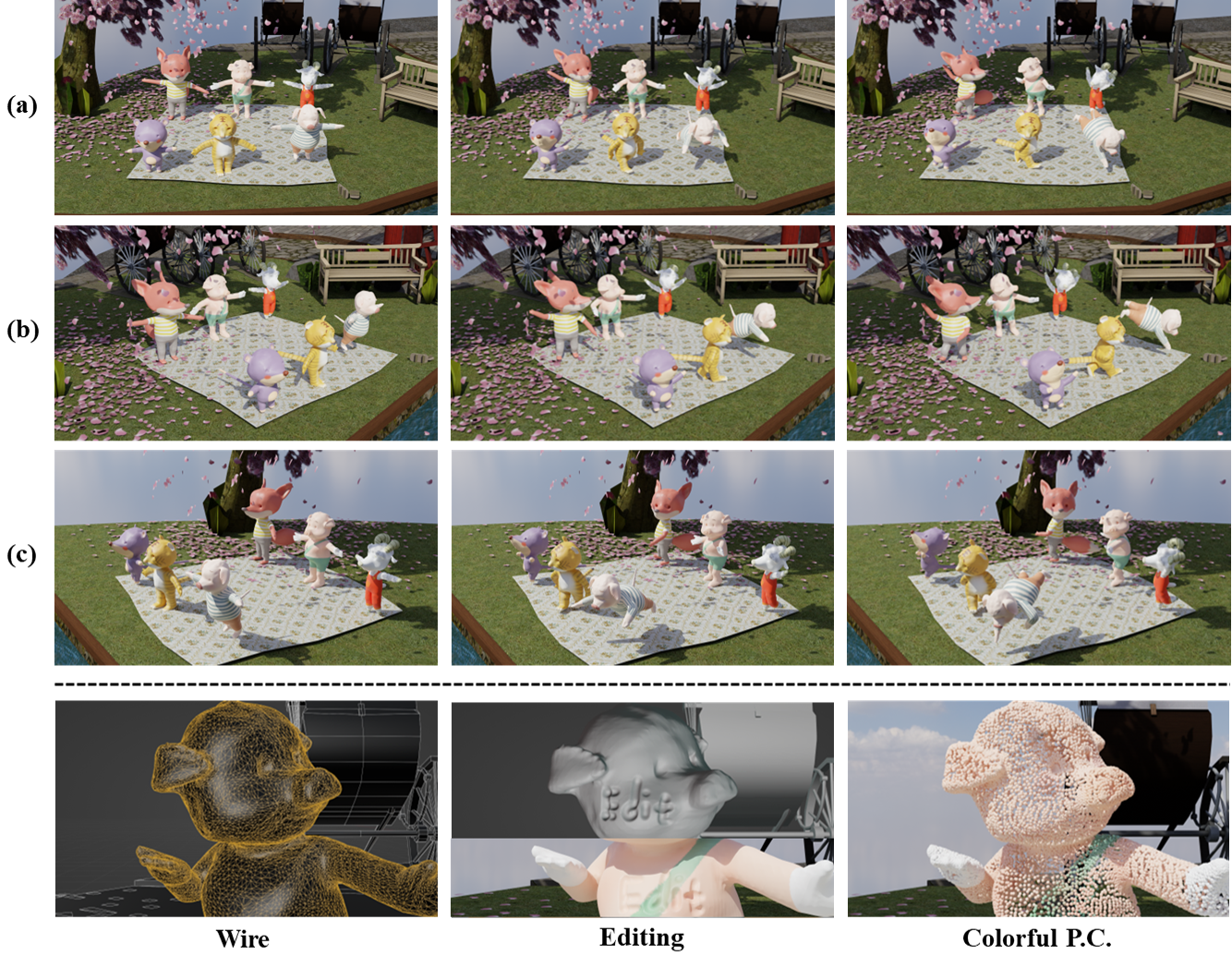}
    \vspace{-3mm}
    \caption{
        Generated volumetric 4D sequence.}
    \label{fig:volumetric-show}
    \vspace{-4mm}
\end{figure}

\begin{figure}[t]
    \centering
    \begin{minipage}[c]{0.50\linewidth}
        \centering
        \includegraphics[width=\linewidth]{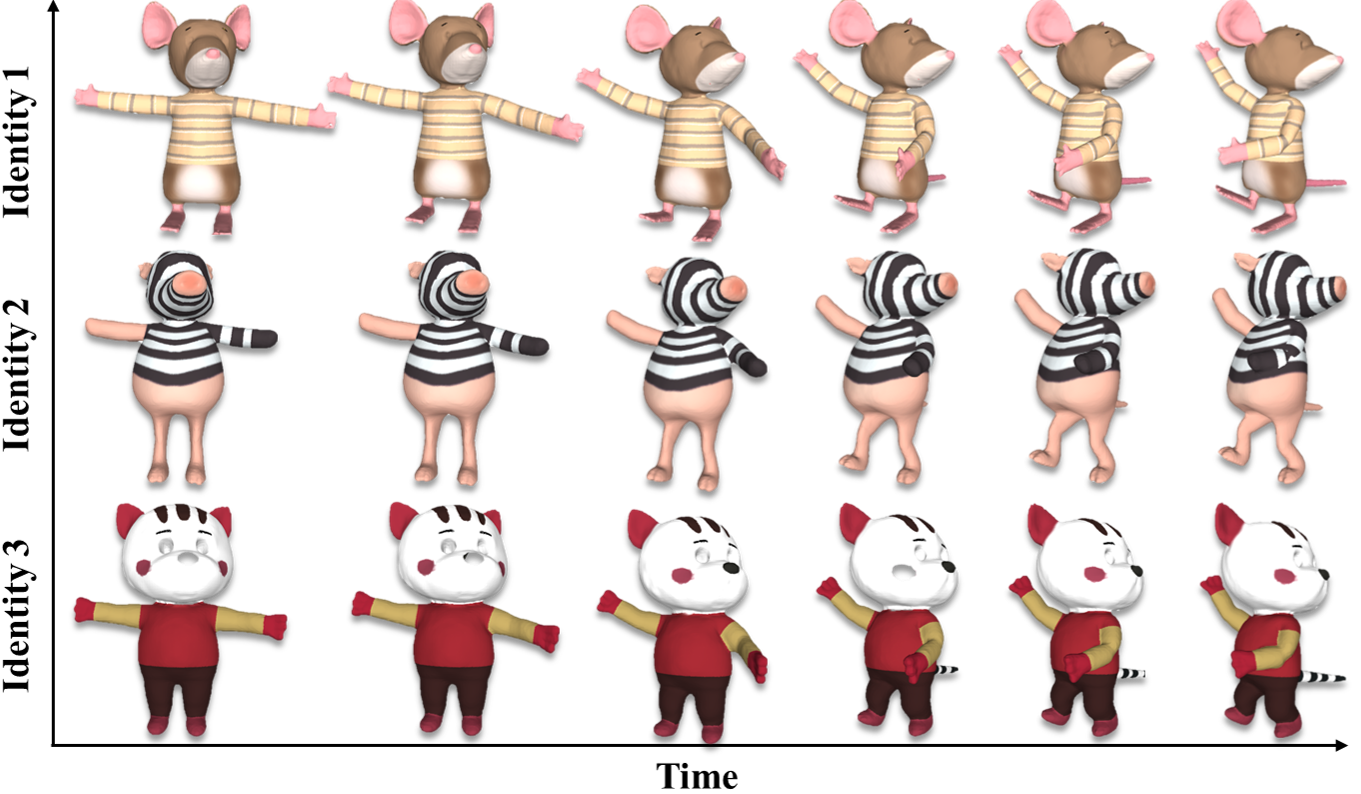}
        \vspace{-10mm}
        \caption{Identity transfer experiments.}
        \label{fig:identity-transfer}
    \end{minipage}
    \begin{minipage}[c]{0.48\linewidth}
        \centering
        \setlength\tabcolsep{1pt}
        \scriptsize
        \begin{tabular}{c | c c c c c c}
            \hline \hline
            \multirow{2}{*}{Method} & \multirow{2}{*}{Speed} & \multirow{2}{*}{Color} & \multicolumn{3}{c}{CLIP-S} \\ 
            \cline{4-6}
             & & & 3DBiCar & Obj.-110k & Obj.-323k \\ \midrule
            DG4D~\cite{ren2023dreamgaussian4d} & 6.5 min & $\checkmark$ & 0.803 & 0.790 & 0.785 \\
            Hyper.~\cite{erkocc2023hyperdiffusion} & 0.54s & $\times$ & - & - & - \\
            STAG4D~\cite{zeng2024stag4d} & 7.5 min & $\checkmark$ & 0.908 & 0.895 & 0.906 \\
            L4GM~\cite{ren2024l4gm} & 3.7s & $\checkmark$ & 0.940 & 0.925 & 0.930 \\
            GenMOJO~\cite{chu2025robust} & 46.3s & $\checkmark$ & 0.909 & 0.911 & 0.903 \\
            \textbf{Ours} & 0.625s & $\checkmark$ & 0.911 & 0.905 & 0.915 \\
            \hline \hline
        \end{tabular}
        \vspace{-2mm}
        \caption{Comparison of 4D generation methods.}
        \label{tab:model-comp}
    \end{minipage}
    \vspace{-2mm}
\end{figure}

\noindent\textbf{Generation via conditions.} 
First, we present our conditional 4D sequence generation results in Fig.~\ref{fig:conditonal-4d}. We show images from different viewing angles and depth maps of the generated sequences to demonstrate the ability to produce plausible, topologically continuous, and visually consistent 4D sequences. Fig.~\ref{fig:conditonal-4d} illustrates that our model effectively learns color and shape information, and generates 4D motion sequences from image and text prompts. Because our method generates the 4D representation $\mathcal{M}$ by balancing the trade-off between 3D shape quality and sequence coherence, which can be accurately decoded into 4D sequences maintaining diversity and alignment with image-text guidance, benefiting from our coherent 3D shape and color modeling. Meanwhile, benefiting from our HCSTA attention module, we explicitly supervise the correlation between shape and color while maintaining spatiotemporal consistency.

\noindent\textbf{Variable-length 4D sequences.} 
To demonstrate the capability of generating arbitrarily long 4D sequences, we create latent 4D representations of the same object performing different actions using varying motion prompts, then apply interpolation to generate intermediate frames. Fig.~\ref{fig:long-seq} shows that our method can produce variable-length 4D sequences through identity transfer and interpolation. This validates the practicality of our approach for generating coherent 4D sequences in stages and interpolations. Because after sufficient training, our matrixized 4D representation forms a continuous and meaningful latent space, allowing variable-length 4D sequences to be constructed via interpolation and our coherent 3D shape and color modeling.

\noindent\textbf{Volumetric property.} 
To demonstrate the volumetric property of our generated results, we create volumetric 4D sequences, place them in a shared scene, and perform free navigation and rendering. Results in Fig.~\ref{fig:volumetric-show} show that our method supports navigation across different views (a), (b), and (c) while maintaining high-quality rendering. Additionally, our approach enables free editing and manipulation of each frame, distinguishing it from other free-form 4D generation models. This is because our coherent 3D shape and color modeling, together with the matrixized 4D representation, constructs volumetric 4D sequences in which each frame is a high-fidelity, editable mesh. Compared to 4D representations based on 3D Gaussian Splatting or video generation, our method offers superior editability and better compatibility with standard computer graphics pipelines, distinguishing it from existing approaches.

\noindent\textbf{Cross Identity transfer.} 
To evaluate the controllability of our method, we apply the same motion prompt to images of different identities, generating 4D sequences with consistent motion but varying identities. Results in Fig.~\ref{fig:identity-transfer} show that our method retains the input's identity and appearance while accurately capturing the motion described in the prompt. This is because our latent matrixized 4D sequence representation effectively decouples shape, color, and motion. With the HCSTA layer, we can explicitly supervise the spatiotemporal information of shape, appearance, and motion, enabling the same motion to be performed across different objects.

\noindent\textbf{Comparison of 4D Generative Models.} We first compare \textit{overall quality and generation speed}. Our model uses diffusion to generate latent 4D sequence representation $\mathcal{M}$ from image and text conditions. First, we compare it with DreamGaussian4D (DG4D)~\cite{ren2023dreamgaussian4d}, HyperDiffusion~\cite{erkocc2023hyperdiffusion}, STAG4D~\cite{zeng2024stag4d}, L4GM~\cite{ren2024l4gm}, and GenMOJO~\cite{chu2025robust} in Tab.~\ref{tab:model-comp} using 3DBiCar and Objaverse 110k and 323k subsets. DG4D supports colored 4D sequences but is slow and memory-intensive. Although STAG4D significantly enhances output quality, this improvement is accompanied by increased computational overhead. HyperDiffusion, while fast, compromises on both color and geometric quality. Although L4GM and 4Diffusion achieve similar generation quality to ours, they require significantly more time and memory. In contrast, our method balances speed and memory efficiency in generation by leveraging matrixized 4D representation.

We also test the \textit{spatiotemporal consistency} and the \textit{alignment} with the input using L4GM and 4Diffusion (video-based) using LPIPS and CLIP-S for overall quality, CLIP-T and FVD for consistency (30s), AfC for the average number of artifacts, and measure max action length and processing time for efficiency in Tab.~\ref{tab:discussion}. 
\begin{table}[t]
    \centering
    \small
    \setlength\tabcolsep{2.7pt}
    \begin{tabular}{c|ccccccc}
    \hline \hline
    Method & LPIPS & CLIP-S & CLIP-T & FVD & AfC & Length & Time \\ \midrule
    Ours & 0.13 & 0.91 & \textbf{0.0096} & \textbf{661.31} & \textbf{0.6} & \textbf{60} & \textbf{1.8s} \\
    L4GM~\cite{ren2024l4gm} & 0.12 & 0.94 & 0.0099 & 691.87 & 2.8 & 14 & 7.2s \\
    4Diff.~\cite{zhang20244diffusion} & 0.12 & 0.92 & 0.0113 & 705.61 & 6.1 & 8 & 600s \\ \hline \hline
    \end{tabular}
    \vspace{-2mm}
    \caption{Comparison of the L4GM and 4Diffusion using various metrics.}
    \label{tab:discussion}
    \vspace{-3mm}
\end{table}
\begin{figure}[t]
    \centering
    \begin{minipage}[c]{0.54\linewidth}
        \centering
        \resizebox{\linewidth}{!}{
            \includegraphics{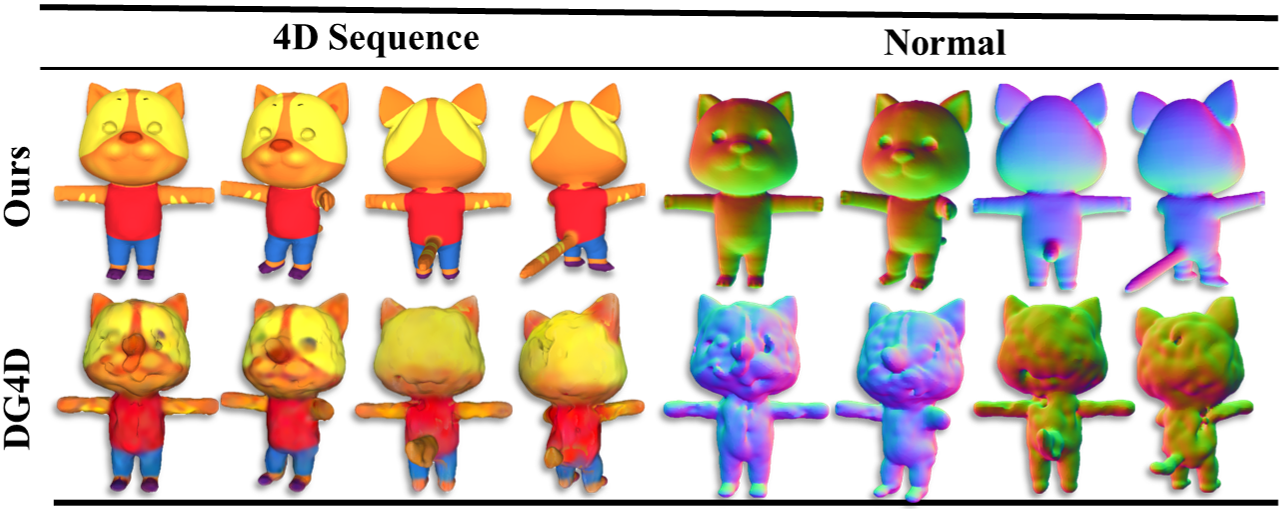}
        }
        \vspace{-8mm}
        \captionof{figure}{Qualitative comparison with DreamGaussian4D.}
        \label{fig:view-4d-comp}
    \end{minipage}
    \begin{minipage}[c]{0.45\linewidth}
        \centering
        \resizebox{\linewidth}{!}{
            \begin{tabular}{c|ccc}
                \hline \hline
                Method & Quality & Spat. Cons. & Alignment\\ \midrule
                Ours (110k) & 4.37 & 4.88 & 4.31 \\
                Ours (323k) & 4.31 & 4.85 & 4.28 \\
                STAG4D (110k) & 4.00 & 4.50 & 4.05 \\
                STAG4D (323k) & 3.95 & 4.45 & 3.98 \\ \hline \hline
            \end{tabular}
        }
        \vspace{-2mm}
        \captionof{table}{User study ratings for generated 4D sequences on 110k and 323k subsets of Objaverse.}
        \label{tab:user-study-ratings}
    \end{minipage}
\end{figure}
Tab.~\ref{tab:discussion} indicates that L4GM’s 3D Gaussian representation excels in appearance quality (LPIPS, CLIP) but is computationally expensive with limited sequence length. Video-based methods offer visual richness but lack geometric guidance, leading to illusions. Our method outperforms in efficiency (time), consistency (CLIP-T, FVD), sequence length, and avoids illusions. This is because our volumetric 4D design explicitly incorporates geometric information and decouples shape from color, enabling explicit supervision of per-frame geometry and appearance. This significantly enhances spatiotemporal consistency and reduces the average number of artifacts.

We conduct more experiments on the 110k subset and 323k of Objaverse in Tab.~\ref{tab:user-study-ratings} to validate the quality at different scales, spatiotemporal consistency, and alignment with conditions through a user study, where participants are asked to rate each example in these three aspects on a scale from 1 to 5. Our model outperforms STAG4D on all metrics. This is because our method maintains better temporal coherence and input alignment by explicitly decoupling shape and color, and employing a staged HCSTA attention module to enhance input consistency and explicitly supervise spatiotemporal coherence.

We further compare the \textit{geometric quality} with 4D generation models based on 3D Gaussian Splatting.
We compare our method with DG4D in Fig.~\ref{fig:view-4d-comp}, providing color images and normal maps from different views to highlight the differences. Due to GPU limitations, DG4D exhibits poor rendering quality outside supervised views. In contrast, our model generates 4D sequences with more complete appearances and smoother normal and depth maps. Our approach outperforms DG4D by balancing memory efficiency with high-quality sequence generation. This is because our method accurately models per-frame geometry and geometry-aware color through coherent 3D shape and color modeling, focusing on accurate shape representation for each identity, whereas DG4D emphasizes the rendering effects of Gaussian primitives without a precise geometry representation, resulting in lower geometric accuracy.

\vspace{-2mm}
\section{Conclusion}
\vspace{-2mm}
We propose an image-text conditioned volumetric 4D sequence generation method balancing 3D shape quality with sequence coherence, while maintaining diversity and alignment with input guidances. Experiments demonstrate our method's efficiency and superior performance. We hope it will be a useful attempt that opens up new possibilities for the representation and generation of complex 4D sequences.

\noindent\textbf{Limitations.} 
The lack of large-scale real-world datasets limits 4D generation performance. Improved 3D/4D representations and the use of generative priors will be key areas for future breakthroughs.

\bibliographystyle{cas-model2-names}
\bibliography{cas-refs}

\end{document}